\documentclass[a4paper,fleqn]{cas-dc}

\usepackage[numbers]{natbib}

\usepackage{amsmath}
\usepackage{amsthm}
\usepackage{amssymb}

\usepackage{algorithm}      
\usepackage{algpseudocode}  

\usepackage{algpseudocode} 
\usepackage{hyperref}  
\usepackage{pgfplots}
\usepackage{float}
\usepackage{booktabs}
\usepackage{multirow}
\usepackage{threeparttable}
\usepackage{subcaption}

\def\tsc#1{\csdef{#1}{\textsc{\lowercase{#1}}\xspace}}
\tsc{WGM}
\tsc{QE}
\tsc{EP}
\tsc{PMS}
\tsc{BEC}
\tsc{DE}

\begin{document}
\renewcommand{\dbltopfraction}{0.95}   
\renewcommand{\dblfloatpagefraction}{0.9}
\renewcommand{\topfraction}{0.95}      
\renewcommand{\bottomfraction}{0.95}   
\renewcommand{\textfraction}{0.05}     
\setcounter{dbltopnumber}{3}           
\setcounter{topnumber}{2}

\let\WriteBookmarks\relax
\def\floatpagepagefraction{1}
\def\textpagefraction{.001}
\shorttitle{Structure-Aware Distributed Backdoor Attacks in Federated Learning}
\shortauthors{Wang Jian et~al.}

\title [mode = title]{Structure-Aware Distributed Backdoor Attacks in Federated Learning}                      

\author[1,2]{Wang Jian}[style=chinese, prefix=Dr, orcid=0000-0001-7511-2910]
\cormark[1]
\fnmark[2]
\ead{jenseWang@outlook.com}
\credit{Conceptualization, Methodology, Writing - Original Draft}

\author[1,3]{Shen Hong}[style=chinese, prefix=Prof]
\fnmark[3]

\credit{Conceptualization, Methodology, Supervision}

\author[1]{Ke Wei}[style=chinese, prefix=Prof]
\fnmark[1]

\credit{Project administration, Supervision}

\author[4]{Liu Xue Hua}[style=chinese, prefix=Prof]

\credit{Data Curation, Resources}

\cortext[1]{Corresponding author: jenseWang@outlook.com}


\affiliation[1]{organization={Faculty of Applied Sciences, Macao Polytechnic University},
	addressline={R. de Luís Gonzaga Gomes},
	city={Macao},
	postcode={999078},
	country={China}}

\affiliation[2]{organization={Faculty of Cyberspace Security, Software Engineering Institute of Guangzhou},
	addressline={No.548 Guangcong South Road, High-tech Industrial Park, Conghua Economic Development Zone},
	city={Guangzhou},
	postcode={501900},
	country={China}}

\affiliation[3]{organization={School of Engineering and Technology, Central Queensland University},
	addressline={Bruce Highway, Norman Gardens},
	city={Rockhampton},
	state={Queensland},
	postcode={4702},
	country={Australia}}

\affiliation[4]{organization={Faculty of  Software and Artificial Intelligence, Software Engineering Institute of Guangzhou},
	addressline={No.548 Guangcong South Road, High-tech Industrial Park, Conghua Economic Development Zone},
	city={Guangzhou},
	postcode={501900},
	country={China}}

\fntext[fn1]{Supported by the Science and Technology Development Fund, Macao SAR (File No. 0015/2023/RIA1)}
\fntext[fn2]{Supported by the Guangdong Provincial Department of Education (Grant No. 2024KTSCX133).}
\fntext[fn3]{Supported by the Queensland Department of Environment and Science Quantum Challenges 2032 Program (Grant No. Q2032001).}

\begin{abstract}
While federated learning protects data privacy, it also makes the model update process more vulnerable to the long-term effects of stealthy perturbations. Existing studies on backdoor attacks in federated learning mainly focus on trigger design or poisoning strategies themselves, and typically assume that identical perturbations exhibit similar propagation and retention behaviors across different model architectures. This assumption overlooks the impact of model structure on perturbation effectiveness. From a structure-aware perspective, this paper systematically analyzes the coupling relationship between model architectures and backdoor perturbations. We introduce two metrics, Structural Responsiveness Score (SRS) and Structural Compatibility Coefficient (SCC), to characterize a model’s overall sensitivity to perturbations and its relative preference for fractal perturbations. Based on these metrics, we construct a structure-aware fractal perturbation injection framework (TFI) to empirically validate the role of architectural properties in the backdoor injection process of federated learning. Experimental results demonstrate that model architecture has a significant influence on the propagation and aggregation behavior of perturbations. Networks with multi-path feature fusion mechanisms are able to amplify and retain fractal perturbations even under low poisoning ratios, whereas in models with low structural compatibility, the effectiveness of such perturbations is substantially constrained. Further analysis reveals a strong correlation between SCC and attack success rate, indicating that SCC can serve as an effective predictor of perturbation survivability. These findings suggest that backdoor behaviors in federated learning are not solely determined by perturbation form or poisoning intensity, but instead critically depend on the joint effects of model architecture and aggregation mechanisms, providing a new analytical perspective for designing targeted defenses at the structural and system levels.
\end{abstract}


\maketitle

\section{Introduction}

In recent years, the widespread adoption of artificial intelligence technologies in privacy-sensitive domains such as medical diagnosis, financial risk control, and industrial manufacturing has made it a central challenge to train high-performance models while preserving data privacy \cite{liu2024recent}. Federated learning provides a feasible paradigm for cross-institutional collaborative modeling by keeping data locally and sharing only model updates, thereby mitigating the privacy leakage risks inherent in traditional centralized training
\cite{yurdem2024federated}. However, the decentralized and open participation mechanisms of federated learning also introduce new security vulnerabilities.

In federated learning settings, the server is typically unable to rigorously audit the data sources or local training processes of participating clients. By controlling or masquerading as only a small number of clients, an adversary can inject malicious updates into the training process and exert a persistent influence on the global model \cite{xie2025model}. Among various threats, backdoor attacks are regarded as one of the most dangerous forms due to their high stealthiness. By implanting samples embedded with specific “triggers” during local training, attackers can cause the aggregated global model to maintain normal performance on benign inputs while consistently producing attacker-specified mispredictions when the trigger condition is met
\cite{byzantinegradientdescentHowBackdoorFederated2020}. Such attacks have been demonstrated to be practically feasible in tasks including image recognition, speech recognition, and text classification.

Existing studies on federated learning backdoor attacks have proposed a variety of attack strategies and trigger design schemes. Early approaches, such as model replacement attacks, can inject backdoors with high intensity, but their anomalous model updates are often easily detected by robust aggregation mechanisms or parameter anomaly detection methods. Subsequently proposed distributed backdoor attacks split a global trigger into multiple sub-triggers and inject them collaboratively across different clients, thereby improving stealthiness, but typically require a higher poisoning ratio to sustain attack effectiveness
\cite{xieDBADISTRIBUTEDBACKDOOR2020}. At the trigger design level, most existing methods rely on explicit patterns, random noise, or frequency-domain perturbations. While effective under specific settings, these approaches still suffer from limitations in cross-model transferability and long-term stealthiness
\cite{tran2018spectral}.

A noteworthy yet insufficiently explored observation is that the effectiveness of a trigger is not determined solely by its geometric or statistical properties, but is also closely related to the structural characteristics of the target model. Prior studies have shown that different neural network architectures exhibit significantly different response pathways to input perturbations. For example, the skip connections in residual networks (ResNet) can amplify the cross-layer propagation of perturbation signals, while the dense feature reuse mechanism in DenseNet helps preserve perturbation features across multiple scales \cite{goodfellow2015explaining}. This suggests that, in backdoor attacks, there may exist a form of “structural compatibility” between perturbation design and model architecture, enabling certain types of perturbations to be more easily amplified and retained in specific architectures. However, most existing federated learning backdoor studies implicitly assume that triggers behave similarly across different model structures, overlooking this structure--perturbation interaction mechanism.

Fractal perturbations, due to their self-similar and multi-scale recursive properties, have recently attracted attention in the generation of adversarial examples and stealthy perturbations \cite{wang2025unveilinghiddenthreatsusing}. In the frequency domain, such perturbations typically exhibit power-law distributions, endowing them with broad-spectrum characteristics and strong statistical stealthiness, which under certain conditions enables them to evade detection methods based on spectral or statistical features. Nevertheless, systematic studies on the relationship between fractal perturbations and structure-aware response pathways in neural networks remain limited. In particular, under federated learning scenarios, it is still unclear whether fractal perturbations can exploit structural properties such as residual connections and feature reuse to achieve more efficient and covert backdoor injection. Motivated by these observations, this paper revisits backdoor attacks in federated learning from a structure-aware perspective and focuses on addressing the following three key questions:
\begin{enumerate}
	\item Do fractal perturbations exhibit differentiated propagation and response behaviors across different model architectures?
	\item Can such structure--perturbation compatibility be leveraged to achieve stable and stealthy backdoor attacks under lower poisoning ratios?
	\item Do these attacks exhibit inherent structural constraints that can inform the design of targeted defense mechanisms?
\end{enumerate}
To this end, we propose a structure-aware compatibility analysis framework, TFI, and introduce two metrics, Structural Response Sensitivity (SRS) and Structural Compatibility Coefficient (SCC), to quantify the impact of model architecture on perturbation propagation behavior. Building upon this analysis, we design a structure--temporal collaborative fractal backdoor attack method and systematically investigate its dependence on model architectures and aggregation mechanisms. The main contributions of this work are summarized as follows:
\begin{enumerate}
	\item From a structure-aware perspective, this paper systematically analyzes the impact of model architectures on the propagation and retention of backdoor perturbations in federated learning, revealing a significant coupling between trigger effectiveness and network structure.
	\item We propose two practical quantitative metrics, Structural Response Sensitivity (SRS) and Structural Compatibility Coefficient (SCC), to characterize a model’s overall sensitivity to input perturbations and its relative compatibility with fractal perturbations.
	\item Based on the above structural analysis, we construct a structure-aware fractal perturbation injection framework as an analytical vehicle to empirically validate the impact of model structural properties on perturbation injection efficiency and stealthiness under limited attack budgets.
	\item Through systematic experiments across diverse model architectures, data scales, and defense mechanisms, we verify the strong correlation between structural compatibility and perturbation survivability, and further provide interpretable defense insights from the perspectives of model architecture and federated aggregation mechanisms.
\end{enumerate}

\section{Related Work}

\subsection{Backdoor Attacks in Federated Learning}

Early studies have shown that attackers can efficiently implant backdoors within a single or a small number of communication rounds via model replacement (MR) attacks. However, such methods typically introduce highly anomalous updates, which are easily detectable by robust aggregation mechanisms through parameter distributions or gradient statistics
\cite{blanchard2017machine,yin2018byzantine}. To improve stealthiness, subsequent work proposed distributed backdoor attacks (DBA), which decompose a complete trigger into multiple sub-triggers injected collaboratively by different clients, thereby reducing the abnormality of any single client update. Due to the weakened perturbation signal at each client, these approaches often require higher poisoning ratios or longer attack durations to maintain stable attack performance. 

More recently, backdoor attack strategies have been extended to more complex federated learning settings. For example, BADFSS introduces backdoor attacks into federated self-supervised learning, demonstrating that even in the absence of explicit label supervision, attackers can still induce abnormal behaviors in downstream tasks through structured perturbations \cite{zhang2024badfss}. In addition, multi-objective and multi-trigger backdoor attacks have been proposed to enhance attack flexibility and persistence. Representative methods such as Dual Model Replacement implant multiple stealthy backdoors within a single model \cite{wang2024dual}. Backdoor attacks against personalized federated learning have also been systematically studied, with results indicating that even when client models differ, shared parameters can still serve as effective carriers for backdoor injection
\cite{ye2024bapfl}.

\subsection{Trigger Design}

Traditional backdoor attacks predominantly rely on explicit triggers, such as fixed pixel patterns or geometric shapes. While easy to implement, these triggers exhibit salient characteristics in both the spatial and frequency domains, making them susceptible to detection
\cite{gu2017badnets,wang2019neural}. To improve stealthiness, researchers have explored covert perturbation-based triggers grounded in frequency-domain or statistical distributions. For instance, the Spectral Backdoor method embeds sparse perturbations in the frequency domain, rendering the trigger nearly imperceptible in the spatial domain. Subsequent studies further reveal that different frequency-band perturbations exhibit significantly different retention behaviors during model training, with certain spectral components being more resistant to suppression
\cite{wei2019sparse}. 

More recently, structured and multi-scale perturbations have attracted increasing attention. Fractal perturbations, owing to their self-similar structure and broad-spectrum frequency distributions, have been shown to possess strong statistical stealthiness in adversarial example generation and covert perturbation design. However, existing studies on fractal perturbations mainly focus on adversarial robustness analysis, and their systematic application to federated learning backdoor attacks—as well as their compatibility with different model architectures—remains insufficiently explored.

\subsection{Model Architecture and Perturbation Response}

A substantial body of research has demonstrated that the propagation behavior of input perturbations within neural networks is closely tied to model architecture. Studies on adversarial examples show that small perturbations can accumulate and amplify across layers in deep networks, thereby significantly influencing prediction outcomes. At the architectural level, residual networks provide low-attenuation cross-layer propagation paths through skip connections, enabling perturbations to survive more easily in deep models
\cite{heDeepResidualLearning2016}. DenseNet further enhances feature reuse through dense connectivity, allowing perturbations to be repeatedly propagated along multiple paths
\cite{Huang_2017_CVPR}. In contrast, in sequentially stacked convolutional networks, perturbation signals often decay progressively with depth. 

In recent years, studies on architectures based on self-attention mechanisms, such as Transformers, have suggested that their global weighting schemes may suppress local structured perturbations to some extent, making certain backdoor or adversarial perturbations difficult to persist
\cite{subramanya2024closer}. Although these works reveal the influence of model structure on perturbation response, most analyses are confined to centralized training or adversarial example scenarios, and have not systematically examined their role in federated learning backdoor attacks.

\subsection{Backdoor Defenses in Federated Learning}

Existing defenses against federated learning backdoor attacks mainly fall into two categories: robust aggregation and detection-based defenses. Robust aggregation methods, such as Krum and Trimmed Mean, mitigate attack impact by suppressing anomalous updates, while differential privacy mechanisms introduce noise during training to limit the influence of individual client updates on the global model
\cite{abadi2016deep}. Detection-based defenses aim to identify potential backdoor behaviors either during training or after model aggregation, including methods based on gradient statistics, parameter distributions, or frequency-domain features. Recent studies further evaluate the effectiveness of these defenses under complex attack settings, such as multi-trigger attacks, frequency-domain perturbations, or Transformer-based architectures, and find that existing methods still exhibit notable limitations when confronting structured or broad-spectrum perturbations
\cite{lee2024detrigger}.

Although existing research on federated learning backdoor attacks has continuously evolved in terms of attack strategies and trigger design, it largely focuses on data- and statistics-level stealthiness. Most studies implicitly assume that triggers exhibit similar effects across different model architectures, overlooking the critical role of model structure in perturbation propagation, retention, and federated aggregation. This work directly addresses this gap by systematically analyzing the propagation characteristics of fractal perturbations in federated learning from a structure-aware perspective, and by proposing a new attack and analysis framework grounded in architectural considerations.

\section{Method}

The success of backdoor attacks in federated learning does not depend solely on the form of the trigger itself, but is also closely related to the structural characteristics of the target model. In particular, in modern deep networks, architectural components such as residual connections and feature reuse provide multi-path propagation channels for input perturbations, making certain perturbations easier to amplify and retain during parameter updates. Owing to their multi-scale and self-similar properties, fractal perturbations naturally exhibit the potential to synergize with such structures. From the perspective of perturbation propagation, this section proposes a structure-aware compatibility analysis framework to characterize how different model architectures respond to perturbations. This framework provides theoretical guidance for client selection and perturbation injection strategies in subsequent attack methods, while also explaining why fractal perturbations exhibit stronger stealthiness and attack efficiency in specific model structures.

\subsection{Hierarchical Response Modeling of Perturbation Propagation}

Let the input sample be $x$, the perturbation be $\delta$, and the deep model $f(\cdot)$ consist of $L$ hierarchical modules. Intuitively, the propagation strength of a perturbation within a network depends on the extent to which it is amplified or attenuated at each layer. To this end, we introduce the concept of hierarchical perturbation response to characterize a model’s sensitivity to input perturbations at different depths. The response of the $l$-th layer to a perturbation is defined as:

\begin{equation}
	R_{l}(\delta)=\left|\frac{\partial f^{(l)}(x+\delta)}{\partial\delta}\right|_{2},
\end{equation}

where $f^{(l)}$ denotes the output of the $l$-th layer. This definition is consistent with gradient-based sensitivity measures commonly used in adversarial example research, and is intended to describe the variation trend of perturbation signals during forward propagation. Based on this, we further define Structural Response Sensitivity (SRS) to measure the overall perceptual capability of a model with respect to perturbations:

\begin{equation}
	\text{SRS}(f,\delta)=\sum_{l=1}^{L}\alpha_{l}\cdot R_{l}(\delta),
\end{equation}

where $\alpha_{l}$ denotes the layer-wise weight, reflecting the relative importance of different layers within the model architecture. In general, deeper layers or layers with residual or dense connections exert greater influence on the final prediction, and therefore are assigned higher weights in the computation of SRS.

A larger SRS value indicates that the model is more sensitive to input perturbations and that perturbation signals are less likely to be suppressed during propagation. Conversely, a lower SRS suggests that perturbations are more easily filtered out as they propagate through the network. This metric provides a unified perspective for analyzing the “amplification capability” of different model architectures with respect to backdoor perturbations.

\subsection{Structural Compatibility Measure for Fractal Perturbations}

The core characteristics of fractal perturbations lie in their multi-scale self-similarity and broad-spectrum distribution in the frequency domain \cite{mandelbrot1982fractal}. Unlike traditional static triggers that concentrate energy in a limited number of frequency bands, fractal perturbations distribute energy across multiple bands simultaneously, making them more amenable to propagation along multi-path structures within neural networks. To characterize the degree to which a model architecture is compatible with fractal perturbations, we introduce Structural Compatibility Coefficient (SCC), which compares a model’s relative response strength to fractal perturbations versus traditional triggers:

\begin{equation}
	\text{SCC}(f)=\frac{\text{SRS}(f,\delta_{\text{fractal}})}{\text{SRS}(f,\delta_{\text{static}})}.
\end{equation}

Intuitively, SCC describes whether a given model structure is more “friendly” to fractal perturbations:
\begin{enumerate}
	\item When $\text{SCC}>1$, the model responds more strongly to fractal perturbations than to traditional triggers, making such perturbations easier to encode into parameter updates;
	\item When $\text{SCC}<1$, the propagation of fractal perturbations is constrained in the given structure, and their attack advantage is difficult to realize.
\end{enumerate}
This metric provides a key insight for attackers: not all clients are equally “valuable” for fractal perturbations, as the model architecture itself determines whether a perturbation can effectively survive.

\subsection{Perturbation Retention Mechanism in Federated Aggregation}

In federated learning, the global model is iteratively updated by aggregating local updates from participating clients. The influence of a client’s uploaded update on the global model depends not only on the update magnitude and participation weight, but also on the structural response characteristics of the model to perturbations~\cite{al2025contribution}. Let the global model after aggregation at round $t$ be:

\begin{equation}
	w^{(t+1)}=w^{(t)}+\sum_{i\in\mathcal{S}_{t}}\gamma_{i}\cdot\Delta w_{i}^{(t)},
\end{equation}

where $\gamma_{i}$ denotes the client weight and $\Delta w_{i}^{(t)}$ represents the local update. We abstract the effective impact of the perturbation introduced by client $i$ on the global model as:

\begin{equation}
	\text{Impact}=\gamma_{i}\cdot g\big(\text{SRS}(f_{i},\delta),,\text{SCC}(f_{i})\big),
\end{equation}

where $g(\cdot)$ denotes the joint effect of SRS and SCC. This formulation indicates that whether a perturbation can survive and accumulate in the global model does not merely depend on the number of malicious clients, but rather on whether the attacker preferentially exploits clients whose model structures are more “sensitive.” Consequently, under realistic scenarios with limited attack budgets, selecting clients with higher SRS and SCC is more effective than blindly increasing the poisoning ratio.

The above analysis demonstrates that the effectiveness of backdoor perturbations in federated learning is strongly dependent on the model structure’s ability to propagate and retain perturbations. Architectures with multi-path propagation properties provide low-attenuation channels for perturbations, enabling multi-scale perturbations to survive more easily during parameter updates and federated aggregation. Due to their broad-spectrum and multi-scale nature, fractal perturbations are more likely to form synergistic interactions with such structures, thereby achieving higher injection efficiency under the same attack budget. Therefore, the success of backdoor attacks is determined not only by poisoning ratios or single-round intensity, but also by the structural compatibility of client models. This observation provides a theoretical foundation for subsequent structure-aware client selection and perturbation injection strategies.

\section{Attack Implementation}

\begin{figure*}[htbp]
	\centering
	\includegraphics[width=0.9\linewidth]{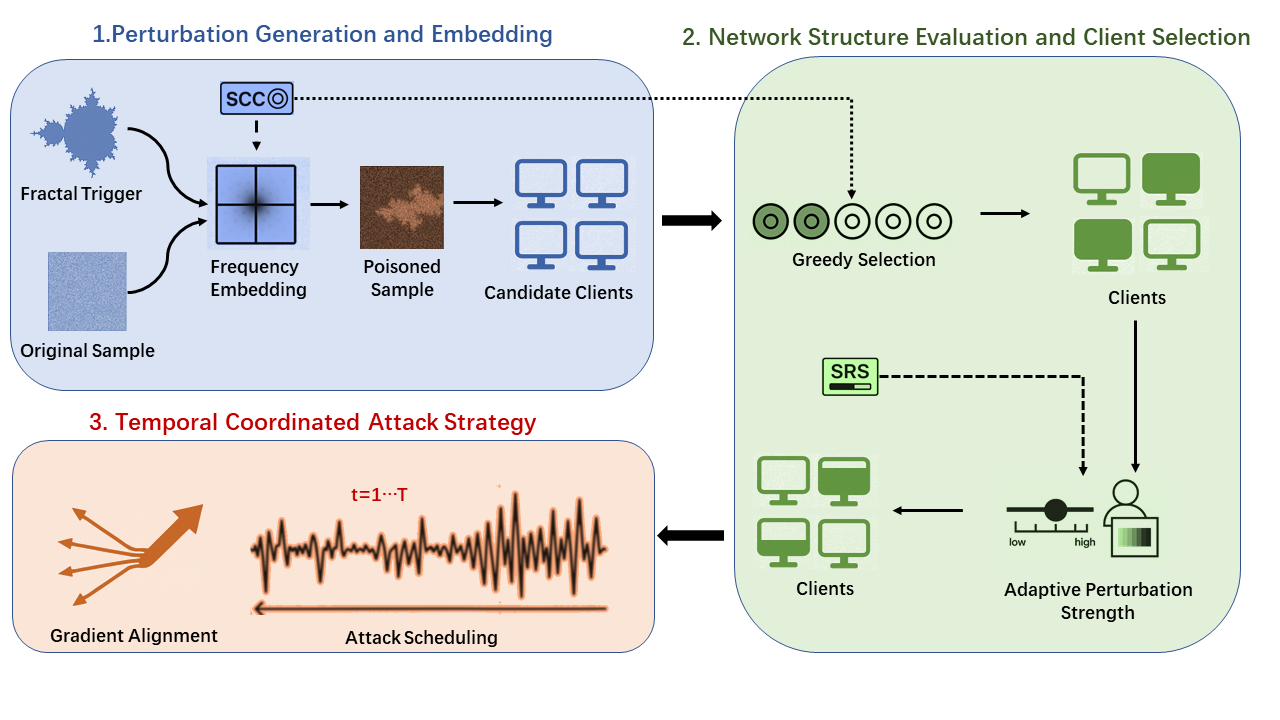}
	\caption{Overview of the TFI backdoor attack framework in federated learning}
	\label{fig:Overview-of-theTFI}
\end{figure*}

Building upon the previously introduced structure-aware compatibility theory, this section presents an executable federated learning backdoor attack method, termed TFI (Structure-aware Fractal Injection). The core objective of TFI is to fully exploit the differentiated responses of model architectures to fractal perturbations under a limited attack budget, thereby writing backdoor behaviors into the global model with higher efficiency and lower detection risk. As illustrated in Figure.~\ref{fig:Overview-of-theTFI}, TFI consists of three synergistic modules: fractal trigger generation and embedding, model structure evaluation and client selection, and a temporally coordinated attack strategy. The implementation details of each module are described below.

\subsection{Fractal Trigger Generation and Embedding}

The design objective of fractal triggers is to construct a structured trigger with multi-scale self-similarity and a broad-spectrum distribution in the frequency domain, thereby avoiding reliance on fixed geometric patterns or single-band energy concentration as in traditional triggers. Compared with static triggers, such perturbations are more likely to propagate along multi-path structures in deep networks and to be retained during parameter updates.

\begin{figure*}[htbp]
	\centering
	\includegraphics[width=0.9\linewidth]{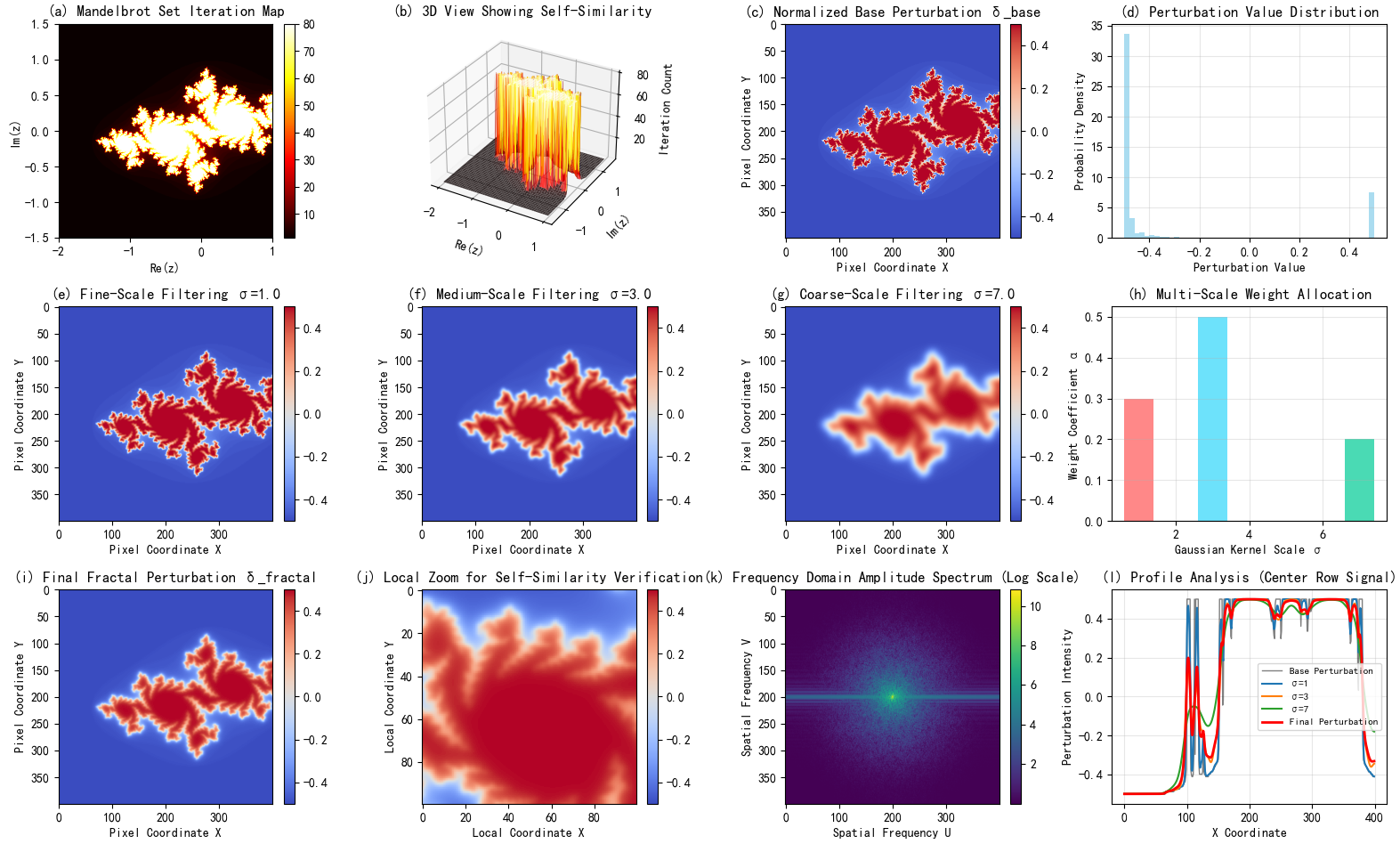}
	\caption{Illustration of fractal perturbation generation and frequency-domain embedding}
	\label{fig:fractal_generation}
\end{figure*}

In practice, we start from a self-similar fractal template $\delta_{\text{base}}$ and generate the final fractal perturbation through multi-scale filtering:
\begin{equation}
	\delta_{\text{fractal}}=\sum_{k=1}^{K}\alpha_{k}\cdot\mathcal{G}{\sigma_{k}}\delta_{\text{base}},
\end{equation}
Figure.~\ref{fig:fractal_generation}
illustrates the overall process from the base template to multi-scale synthesis, followed by frequency-domain embedding into the original samples. As shown, the perturbation does not rely on a fixed geometric shape in the spatial domain, but exhibits self-similar structures across multiple scales; in the frequency domain, its energy distribution is more dispersed, which helps improve stealthiness and survivability during model training and aggregation.

Here, $\mathcal{G}_{\sigma_{k}}$ denotes a Gaussian kernel with scale $\sigma_{k}$, and $\alpha_{k}$ represents the multi-scale weights. This process is performed offline only once and is decoupled from specific training samples. To effectively inject backdoor signals while maintaining stealthiness, we adopt a frequency-domain hybrid embedding strategy. Specifically, we transform the original sample $x$ and the fractal perturbation $\delta_{\text{fractal}}$ into the frequency domain, obtaining $X(\omega)$ and $\Delta(\omega)$, respectively, and then perform weighted superposition:
\begin{equation}
	X_{\text{poison}}(\omega);=;X(\omega);+;\beta_{i}(\omega)\cdot\Delta(\omega).
\end{equation}

\begin{figure*}[htbp]
	\centering
	\includegraphics[width=0.75\linewidth]{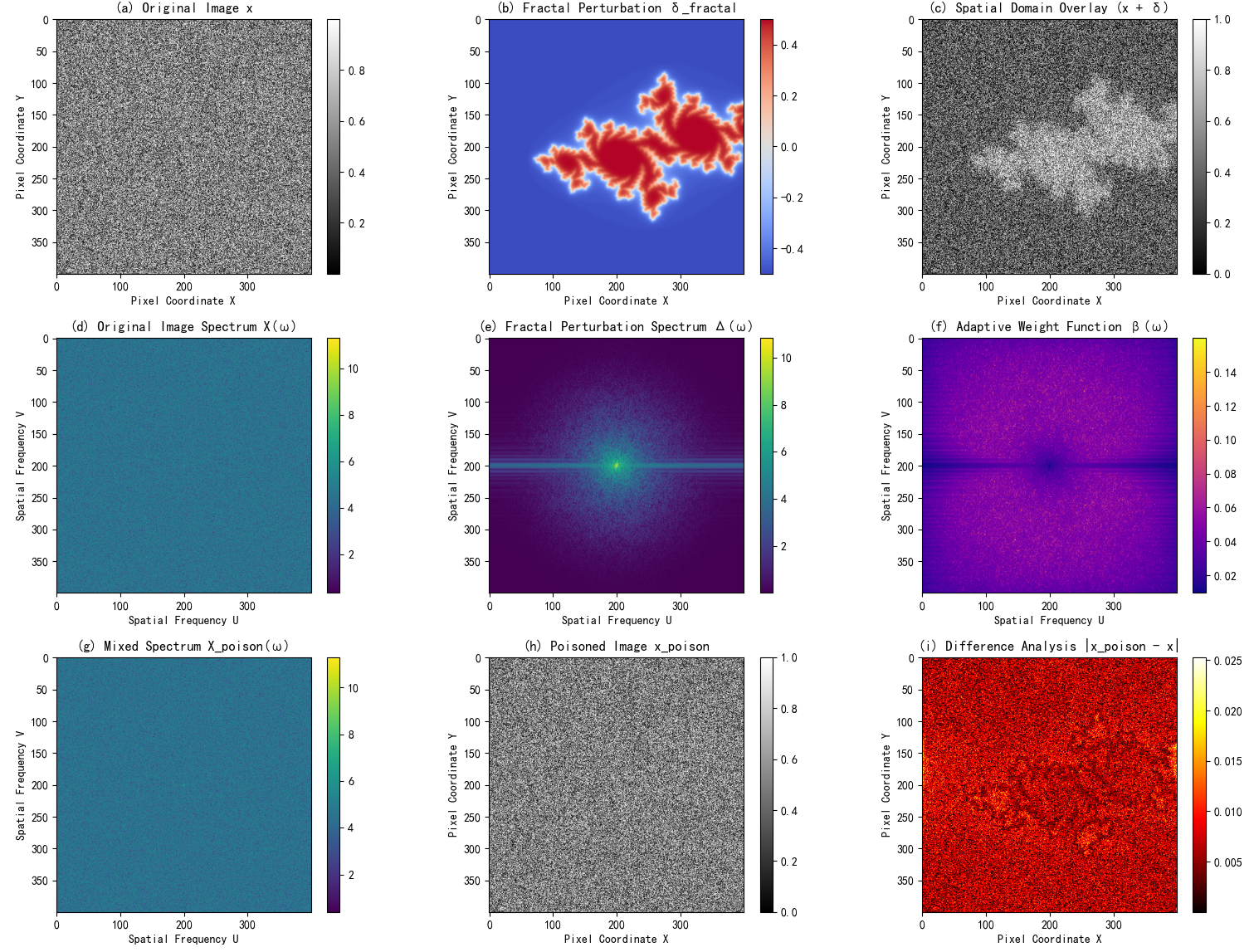}
	\caption{Multi-scale generation of fractal perturbations and their frequency-domain}
	\label{fig:Multi-scale generation of fractal perturbations and their frequency-domain}
\end{figure*}

Fig.~\ref{fig:Multi-scale generation of fractal perturbations and their frequency-domain}
shows the complete pipeline from base template to multi-scale synthesis and frequency-domain embedding, highlighting that the perturbation avoids fixed spatial patterns while exhibiting a more dispersed energy distribution in the frequency domain.

The embedding weight $\beta_{i}(\omega)$ is adaptively adjusted according to the client’s structural compatibility coefficient (SCC):
\begin{equation}
	\beta(\omega)=\epsilon_{i}^{\text{base}}\cdot\mathrm{SCC}_{i}^{\gamma}\cdot w(\omega),
\end{equation}

where $\epsilon_{i}^{\text{base}}$ is the client-level baseline perturbation strength, $\gamma\in(0,1)$ is a sublinear amplification exponent, and $w(\omega)$ is a smooth frequency-domain window function used to suppress high-frequency artifacts. This design aligns perturbation embedding with model structural properties, achieving higher injection efficiency on structurally favorable clients while controlling overall exposure risk.

\subsection{Model Structure Evaluation and Client Selection}

The survivability of fractal triggers in federated aggregation critically depends on the client models’ responses to perturbations. Although clients often share the same model topology, differences in local data distributions, training states, and potential personalization modules can lead to substantial variation in perturbation responses~\cite{liu2025mitigating}. Therefore, under a limited attack budget, prioritizing structurally favorable clients is key to efficient attacks~\cite{zhai2024adaptive}. We adopt an online gradient-response-based estimation method to approximate each client’s SRS and SCC. The server sends a small probe dataset $\mathcal{D}_{\text{probe}}$ to clients and computes gradient norms under clean samples, fractal-perturbed samples, and static-perturbed samples. The estimated SRS of client $i$ is given by:

\begin{equation}
	\widehat{\mathrm{SRS}}(f_{i})=\frac{1}{m}\sum_{j=1}^{m}\Big(|\nabla_{\theta}\mathcal{L}(f_{i}(x_{j}+\delta_{\text{fractal}}))|-|\nabla_{\theta}\mathcal{L}(f_{i}(x_{j}))|\Big),
\end{equation}

and the corresponding SCC estimate is:

\begin{equation}
	\widehat{\mathrm{SCC}}(f_{i})=\frac{\widehat{\mathrm{SRS}}(f_{i})}{\frac{1}{m}\sum_{j=1}^{m}\big(|\nabla_{\theta}\mathcal{L}(f_{i}(x_{j}+\delta_{\text{static}}))||\nabla_{\theta}\mathcal{L}(f_{i}(x_{j}))|\big)}.
\end{equation}

If $\widehat{\mathrm{SCC}}(f_{i})>1$, the client model exhibits a relative preference for fractal perturbations. Based on this, we define the attack value of each client as
$V_{i}=\gamma_{i}\cdot\widehat{\mathrm{SCC}}(f_{i}),$
 where $\gamma_{i}$ is the client’s aggregation weight. Under budget constraints (e.g., maximum number of malicious clients or total weight), the attacker selects the set $\mathcal{A}$ with the highest $V_{i}$ values using a greedy strategy, corresponding to the optimal client selection criterion derived in Chapter~3. To further enhance stealthiness, the baseline perturbation strength is dynamically adjusted according to structural sensitivity:
\begin{equation}
	\epsilon_{i}^{\text{base}}=\epsilon_{0}\cdot\min!\left(2,\frac{\mathrm{SRS}_{\text{ref}}}{\widehat{\mathrm{SRS}}(f_{i})}\right),
\end{equation}

with clipping applied to avoid excessive perturbations on highly sensitive clients.

\subsection{Temporally Coordinated Attack Strategy}

In federated learning, backdoor injection typically requires gradual accumulation over multiple training rounds. Aggressive early attacks may trigger performance anomalies and detection, while persistently weak attacks may fail to accumulate~\cite{ren2024shadow}. To address this trade-off, we design a temporally coordinated attack strategy that schedules attack behavior over time. The global attack intensity is controlled by:
\begin{equation}
	I(t)=I_{\max}\cdot\big(1-e^{-\lambda t}\big),
\end{equation}

which increases slowly in early training and strengthens in later stages, balancing efficiency and stealthiness. At round $t$, the actual perturbation strength for client $i$ is set as:
\begin{equation}
	\epsilon_{i}^{(t)}=\epsilon_{i}^{\text{base}}\cdot\frac{V_{i}}{\bar{V}}\cdot\frac{I(t)}{|\mathcal{S}_{\text{attack}}^{(t)}|},
\end{equation}

where $\bar{V}$ denotes the average client value and $|\mathcal{S}_{\text{attack}}^{(t)}|$ is the number of attacking clients at round $t$. This design ensures coordinated multi-client injection under a controlled total attack intensity.

\subsection{Algorithm Description and Complexity Analysis}

\begin{algorithm}[t]
	\caption{TFI: Structure-Aware Fractal Injection Strategy}
	\label{alg:tfi}
	\begin{algorithmic}[1]
		\Require Initial global model $w^{(0)}$; client set $\mathcal{C}$; probe dataset $\mathcal{D}_{\mathrm{probe}}$; attack budget $\mathcal{B}$; total communication rounds $T$
		\Ensure Compromised global model $w^{(T)}$
		
		\State Construct a multi-scale fractal perturbation pattern $\delta_{\mathrm{fractal}}$ in the frequency domain
		
		\For{each client $i \in \mathcal{C}$}
		\State Evaluate structure-related sensitivity $\widehat{\mathrm{SRS}}(f_i)$ using $\mathcal{D}_{\mathrm{probe}}$
		\State Estimate structural coupling capability $\widehat{\mathrm{SCC}}(f_i)$ via response consistency analysis
		\State Compute client utility score $V_i = \gamma_i \cdot \widehat{\mathrm{SCC}}(f_i)$
		\EndFor
		
		\State Select malicious client subset $\mathcal{A} \subseteq \mathcal{C}$ such that $|\mathcal{A}| \leq \mathcal{B}$
		
		\For{$t = 1$ to $T$}
		\State Compute round-wise global injection intensity $I(t)$
		
		\For{each client $i \in \mathcal{A}$ participating at round $t$}
		\State Compute adaptive perturbation magnitude
		\[
		\epsilon_i^{(t)} = \Phi\!\left(\widehat{\mathrm{SRS}}(f_i),\, V_i,\, I(t)\right)
		\]
		\State Inject $\delta_{\mathrm{fractal}}$ into local training samples via frequency-domain mixing
		\State Perform local optimization and obtain update $\Delta w_i^{(t)}$
		\EndFor
		
		\State Aggregate updates and update global model
		\[
		w^{(t+1)} \leftarrow \mathrm{Aggregate}\left(\{\Delta w_i^{(t)}\}\right)
		\]
		\EndFor
		
	\end{algorithmic}
\end{algorithm}

In terms of complexity, TFI introduces only constant-factor overhead compared to standard FedAvg. The additional computation mainly stems from one-time structural evaluation and lightweight frequency-domain operations, whose cost is negligible relative to model training. Therefore, the method exhibits good scalability in both time and space.

\section{Attack Feasibility Conditions and Defense Insights}

This section formally analyzes the feasibility conditions of the TFI attack from the perspective of federated aggregation, aiming to characterize under what conditions structure-aware fractal backdoors can persistently accumulate in the global model. We abstract the attack process as the effective contribution of perturbations during aggregation and derive necessary inequality conditions for attack success. Based on this analysis, we further discuss how minimal interventions can break these conditions, yielding direct and interpretable defense insights.

\subsection{Attack Mechanism Analysis}

In federated learning, the effective impact of malicious perturbations on the global model at round $t$ can be abstracted as their net contribution during aggregation. Combining the structure-aware analysis in Chapter~3, this contribution can be expressed as:

\begin{equation}
	\Delta w_{\text{adv}}^{(t)}\propto\sum_{i\in\mathcal{A}_{t}}\gamma_{i}\cdot\mathrm{SRS}(f_{i},\delta)\cdot\mathrm{SCC}(f_{i}),
\end{equation}

where $\gamma_{i}$ is the client weight, and $\mathrm{SRS}$ and $\mathrm{SCC}$ characterize the response strength and relative compatibility of the model structure to perturbations, respectively. Backdoor signals can accumulate across training rounds and eventually form stable behaviors when their cumulative effect exceeds benign update fluctuations and system noise, i.e.,

\begin{equation}
	\sum_{t=1}^{T}\Delta w_{\text{adv}}^{(t)}>\sum_{t=1}^{T}\bigl(\Delta w_{\text{benign}}^{(t)}+\xi^{(t)}\bigr),
\end{equation}

where $\xi^{(t)}$ denotes effective noise introduced by differential privacy, clipping, or robust aggregation. When the model structure fails to provide low-attenuation propagation paths, perturbations lack temporal statistical consistency, or aggregation noise dominates the update scale, this inequality no longer holds and the attack signal is naturally suppressed. This condition delineates the structural, statistical, and system-level boundaries for the feasibility of the TFI attack.

\subsection{Defense Insight}

The above analysis indicates that defending against structure-aware fractal backdoor attacks does not require precise trigger identification or explicit recovery of a clean model. Any intervention that systematically disrupts the imbalance in the above inequality can effectively weaken attack feasibility. Reducing multi-path propagation capacity or feature reuse in models directly decreases $\mathrm{SRS}$ and $\mathrm{SCC}$, limiting perturbation contributions during local training and updates; introducing temporal decorrelation or randomization mechanisms disrupts cross-round accumulation; and increasing aggregation noise strength, clipping thresholds, or robustness constraints amplifies the noise term $\xi^{(t)}$, submerging perturbation effects within benign updates. The shared objective of these defenses is not to explicitly detect backdoors, but to push the federated learning system into a parameter regime where structure-aware fractal perturbations cannot persistently accumulate during aggregation.

\section{Experiments}

This section conducts systematic experiments to validate the effectiveness and applicability boundaries of the proposed structure-aware analysis framework and the TFI attack method. In addition to demonstrating the improvement of TFI over existing methods in terms of attack success rate, the experiments aim to answer three key questions: (1) whether the attack efficiency of fractal perturbations is highly correlated with model structural characteristics; (2) whether structural compatibility (SCC) can effectively predict attack performance; and (3) whether the attack degrades as theoretically expected when the structural, statistical, or aggregation conditions required for attack success are disrupted.

\subsection{Experimental Setup}

The experimental design focuses on the impact of model architecture on perturbation propagation and retention, rather than merely comparing absolute attack success rates~\cite{nguyen2024backdoor}. To this end, we conduct comparative analyses across datasets of different scales, model architectures with pronounced structural differences, and multiple defense settings. Two image classification datasets are used: CIFAR-10~\cite{krizhevsky2009learning}and ImageNet-100~\cite{russakovsky2015imagenet}. CIFAR-10 contains 10 classes and 60,000 $32\times32$ color images. To improve experimental efficiency and support multi-round comparisons, we randomly sample 50\% of the dataset and re-split it into 25,000 training samples and 5,000 test samples. ImageNet-100 consists of the first 100 classes of ImageNet-1k, containing approximately 130,000 training samples and 5,000 validation samples, with all inputs resized to $224\times224$. This dataset is used to evaluate the generalization performance of attack methods under large-scale and high-complexity tasks.

In terms of model selection, we adopt neural network architectures with clearly distinct structural paradigms to analyze the relationship between model structure and perturbation propagation. Specifically, ResNet-18 and ResNet-50 represent multi-path architectures with residual connections; DenseNet-121 represents architectures with dense feature reuse; VGG-16 represents traditional sequential convolutional stacking~\cite{simonyan2014very}; and ViT-Base represents global modeling based on self-attention mechanisms~\cite{dosovitskiy2020image}. These structural differences provide sufficient contrast for subsequent quantitative analysis of the relationship between structural compatibility (SCC) and attack efficiency. Table~\ref{tab:Configuration of data set and model structure} summarizes the datasets and model configurations used in the experiments.

\begin{table*}
	\begin{tabular}{|c|c|c|c|}
		\hline
		Dataset & Model & Structural Feature & Input Size\tabularnewline
		\hline
		\hline
		CIFAR-10 & ResNet-18 & Residual connections & $32\times32$\tabularnewline
		\hline
		CIFAR-10 & DenseNet-121 & Dense connections & $32\times32$\tabularnewline
		\hline
		CIFAR-10 & VGG-16 & Sequential convolution & $32\times32$\tabularnewline
		\hline
		CIFAR-10 & ViT-Base & Self-attention & $32\times32$\tabularnewline
		\hline
		ImageNet-100 & ResNet-50 & Residual connections & $224\times224$\tabularnewline
		\hline
		ImageNet-100 & DenseNet-121 & Dense connections & $224\times224$\tabularnewline
		\hline
		ImageNet-100 & VGG-16 & Sequential convolution & $224\times224$\tabularnewline
		\hline
		ImageNet-100 & ViT-Base & Self-attention & $224\times224$\tabularnewline
		\hline
	\end{tabular}
	\caption{\label{tab:Configuration of data set and model structure}Configuration of datasets and model architectures}
\end{table*}

All experiments are conducted in a simulated federated learning environment with 100 clients. Training data are partitioned in a non-IID manner following a Dirichlet distribution with concentration parameter $\alpha=0.5$, reflecting realistic data heterogeneity~\cite{tianComprehensiveSurveyPoisoning2023a}. In each communication round, 10\% of clients are randomly selected to participate in local training, and FedAvg is used for model aggregation. Local training employs stochastic gradient descent (SGD) with momentum 0.9 and weight decay $5\times10^{-4}$. Each client performs 5 local training epochs with a batch size of 32. The global learning rate is initialized at 0.1 and decayed at 50\% and 75\% of the total training rounds.

To comprehensively evaluate the behavior of TFI under different conditions, we compare it with three representative federated backdoor attack methods: model replacement (MR), distributed backdoor attack (DBA), and label poisoning (LP). MR represents a high-intensity but low-stealth attack, DBA improves statistical stealthiness through multi-client collaboration, and LP represents a data-level poisoning approach without explicit input perturbations. The differences between these methods and TFI in terms of attack mechanisms and perturbation forms help isolate the unique role of fractal perturbations from a structure-aware perspective.

For defense evaluation, we focus on attack survivability under typical robust aggregation and detection mechanisms, including the Krum algorithm based on statistical consistency, differential privacy (DP) mechanisms that limit the influence of individual client updates by injecting noise, and the Spectral Signatures method for frequency-domain backdoor detection. These defenses impose constraints at the aggregation, noise, and spectral levels, respectively, providing an experimental basis for analyzing how attack feasibility conditions are disrupted.

Multiple metrics are used to quantitatively evaluate attack effectiveness and stealthiness. Main task accuracy (MTA) measures model performance on clean test data; attack success rate (ASR) evaluates the stability of producing attacker-specified predictions under trigger conditions; update similarity characterizes the statistical proximity between malicious and benign client updates in parameter space; and attack retention rate measures the degradation of attack effectiveness under defense mechanisms relative to the no-defense setting.

\subsection{ASR and MTA under Fixed Poisoning Ratio}

This subsection compares attack behavior across different model architectures under a fixed poisoning ratio. The poisoning ratio is set to 10\% for all experiments, and both ASR and MTA are evaluated to analyze the trade-off between attack effectiveness and performance degradation. Fixing the poisoning ratio effectively isolates the attack budget factor, allowing the experimental results to more directly reflect the influence of model structure on perturbation propagation, accumulation, and aggregation behavior. Under this setting, persistent differences in attack performance can be reasonably attributed to structural responses to fractal perturbations rather than differences in attack intensity.

\begin{figure*}[t]
	\centering
	\includegraphics[width=0.8\linewidth]{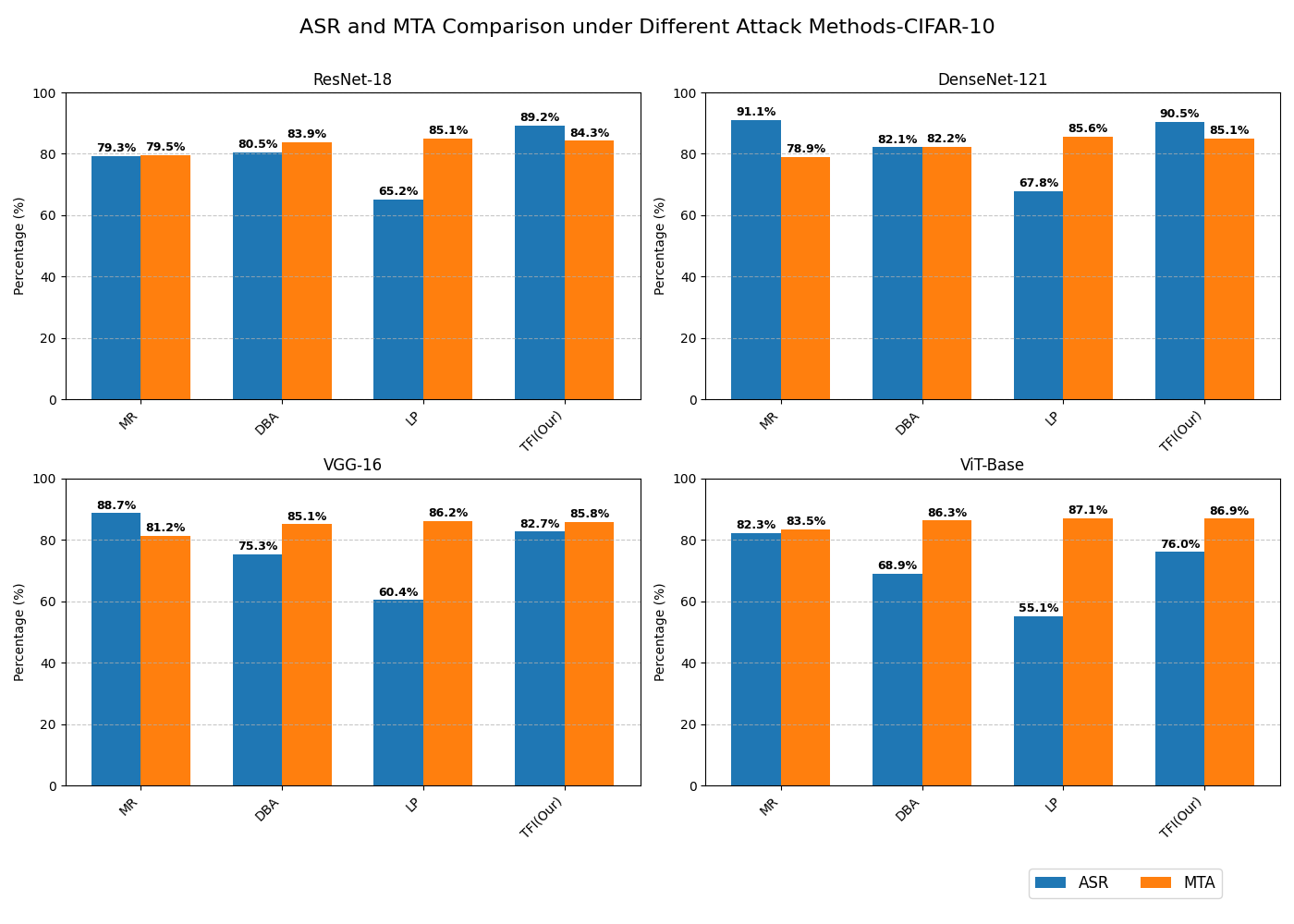}
	\caption{\label{fig:ASR-and-MTA-CIFAR10}ASR and MTA comparison under fixed poisoning ratio on CIFAR-10}
\end{figure*}

Figure~\ref{fig:ASR-and-MTA-CIFAR10} presents the comparison of ASR and MTA across different model architectures on CIFAR-10 under a 10\% poisoning ratio. The results show that in multi-path architectures such as ResNet-18 and DenseNet-121, TFI achieves significantly higher ASR while maintaining nearly unchanged main task performance. This indicates that in architectures with high SCC, fractal perturbations are more easily absorbed and encoded into parameter updates without causing noticeable degradation of the main task. In contrast, in architectures such as VGG-16 and ViT-Base, the ASR of TFI decreases markedly, and its advantage over multi-path structures is substantially reduced. In particular, on ViT-Base, although the main task accuracy remains high, the ASR reaches only 76.0\%, significantly lower than that observed in residual and densely connected networks. These results further confirm the decisive role of model structure in attack effectiveness.

\begin{figure*}[htbp]
	\centering
	\includegraphics[width=0.8\linewidth]{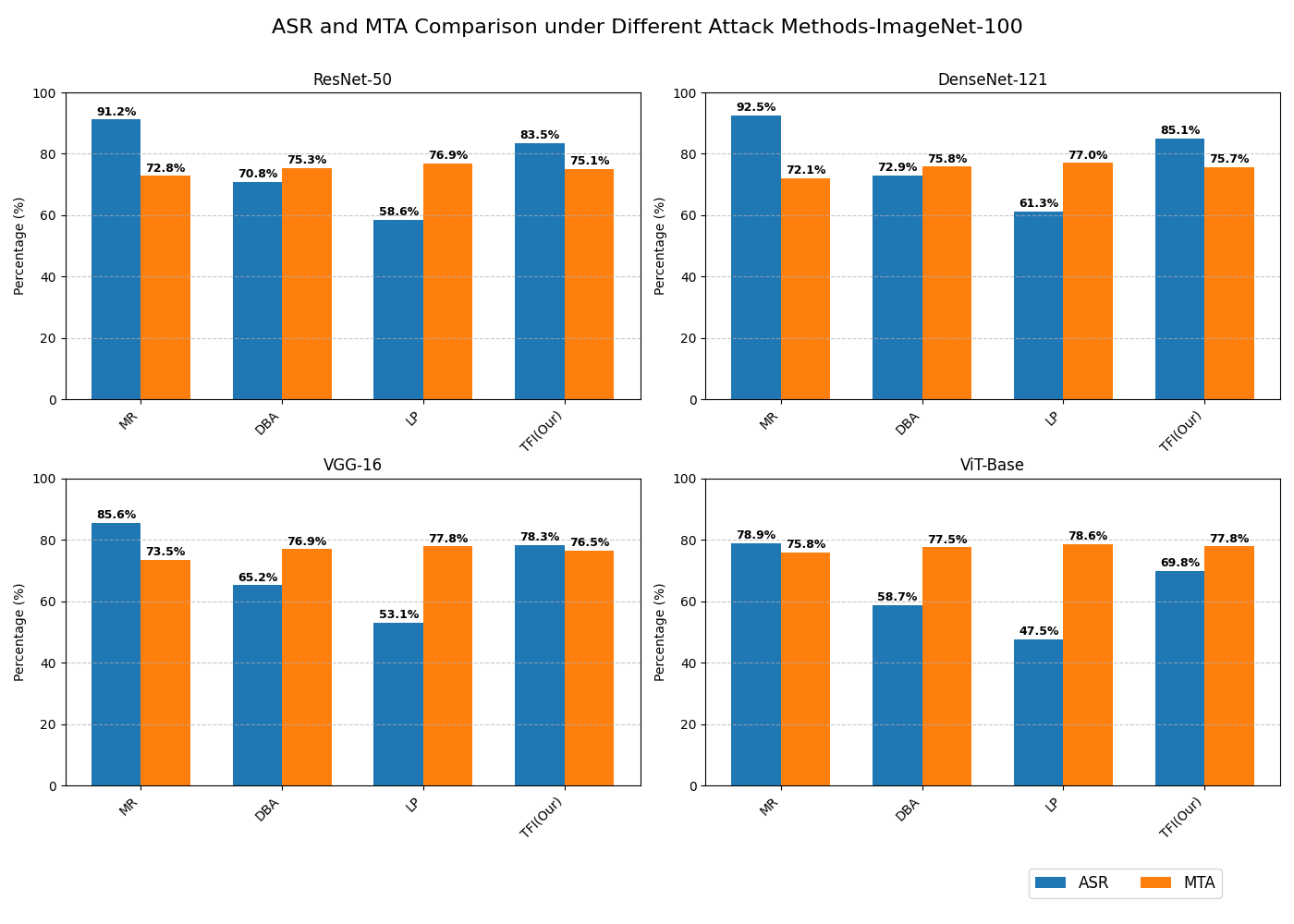}
	\caption{\label{fig:ASR-and-MTA-ImageNet}ASR and MTA comparison under fixed poisoning ratio on ImageNet-100}
\end{figure*}

To verify the consistency of these observations on larger-scale datasets, we conduct the same comparison on ImageNet-100. As shown in Fig.~\ref{fig:ASR-and-MTA-ImageNet}, the overall trend remains consistent with CIFAR-10. In multi-path architectures, TFI maintains high ASR under a fixed attack budget, whereas its effectiveness is significantly constrained in models with lower structural compatibility. This indicates that increasing model scale does not diminish the dominant role of structural factors in attack behavior.

Taken together, these results indicate that under a fixed poisoning ratio, differences in attack success rate are primarily driven by differences in model structure with respect to perturbation propagation and retention, rather than by the inherent strength of the attack method. Meanwhile, TFI induces only limited degradation in main task performance across most architectures, demonstrating strong statistical stealthiness. These findings further validate that when model structures fail to effectively amplify or retain fractal perturbations, attack performance degrades significantly under a fixed budget, consistent with the theoretical analysis of structural compatibility thresholds.

\subsection{Stealthiness and Robust Aggregation}

In the previous subsections, we analyzed the impact of model structure on TFI primarily from the perspectives of attack efficiency and cost. However, in practical federated learning systems, long-term attack effectiveness depends not only on initial injection efficiency but also on perturbation survivability under multi-round aggregation and defense mechanisms. This subsection systematically analyzes the survivability of fractal perturbations during federated training from the perspectives of stealthiness and robust aggregation~\cite{nguyen2021flame}.

\begin{figure*}[htbp]
	\centering
	\includegraphics[width=1\textwidth]{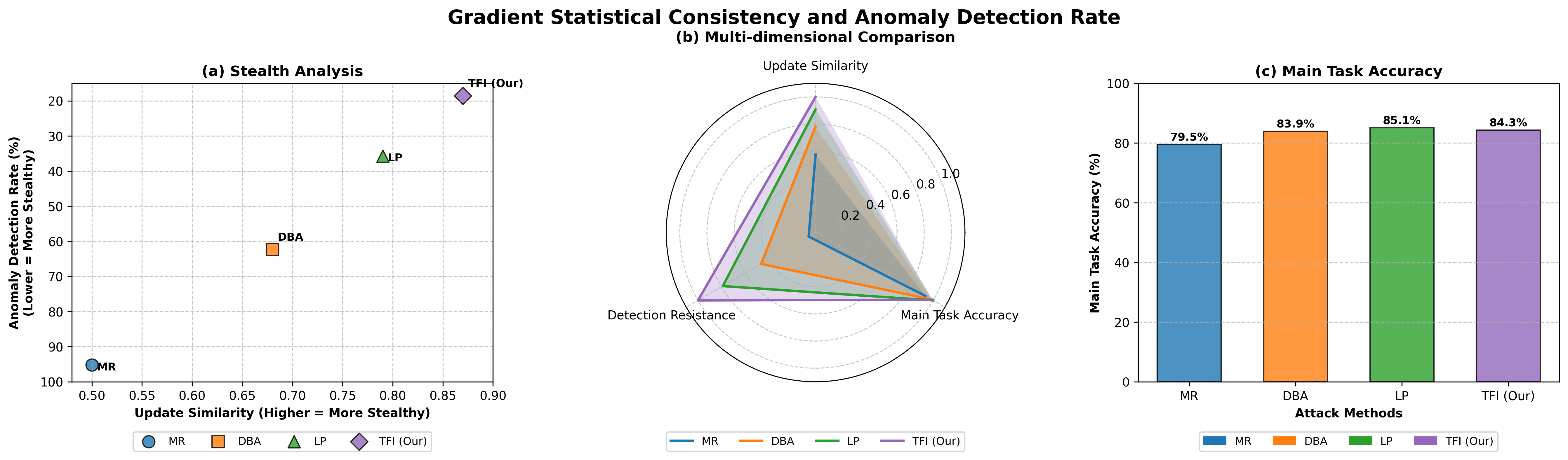}
	\caption{\label{fig:Gradient-Statistical-Consistency}Gradient statistical consistency and anomaly detection rate}
\end{figure*}

\textbf{Gradient statistical consistency.} We measure the cosine similarity between malicious and benign client updates to quantify the statistical consistency of attack updates. Higher similarity indicates greater difficulty for gradient-based anomaly detection methods. Figure~\ref{fig:Gradient-Statistical-Consistency} shows the update similarity and corresponding anomaly detection rates for different attack methods on CIFAR-10 with the ResNet-18 architecture.

The results show that TFI achieves the highest update similarity (0.87) among all compared methods and significantly reduces the anomaly detection rate (18.5\%). In contrast, MR and DBA produce updates that deviate more substantially from benign updates in terms of statistical characteristics, resulting in much higher detection rates. LP exhibits intermediate performance but still falls short of TFI in terms of stealthiness. These results indicate that the fractal perturbations introduced by TFI integrate more naturally into normal gradient distributions during federated aggregation, yielding superior statistical stealthiness.

Moreover, TFI achieves the lowest detection risk while maintaining high main task accuracy (84.3\%), demonstrating a better balance between stealthiness and model performance. These results confirm that TFI exhibits stronger perturbation survivability under robust aggregation settings.

\begin{figure*}[htbp]
	\centering
	\includegraphics[width=0.9\linewidth]{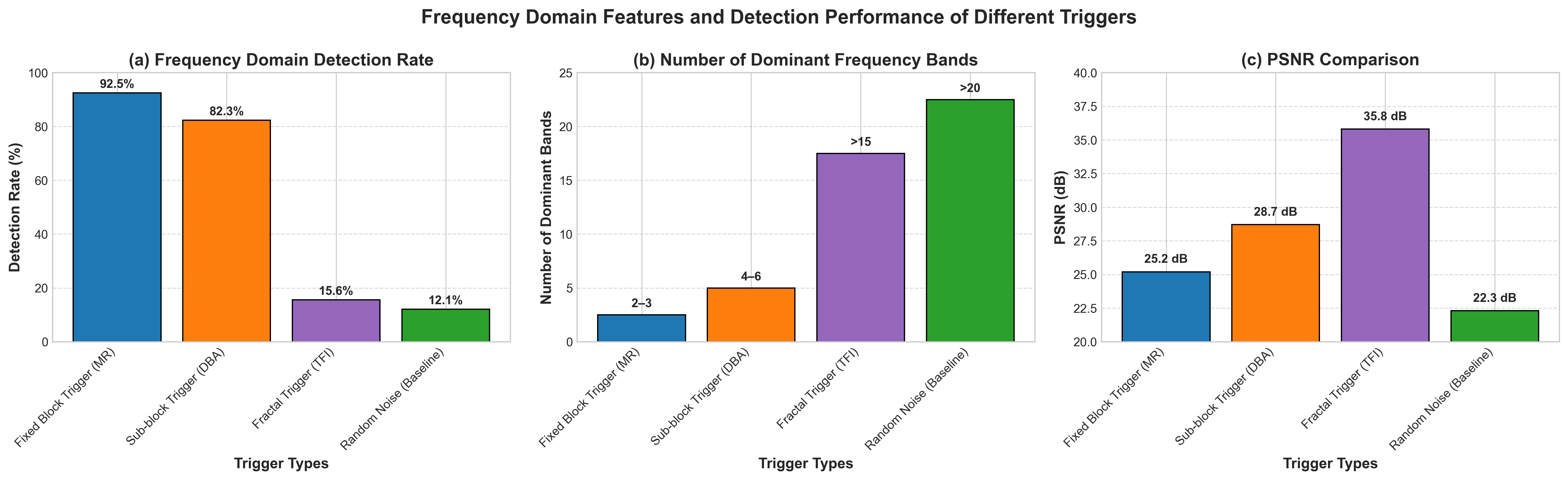}
	\caption{\label{fig:Frequency-Domain-Features}Frequency-domain features and detection performance of different triggers}
\end{figure*}

\textbf{Frequency-domain exposure risk.} In addition to gradient statistics, frequency-domain analysis is a commonly used tool for detecting backdoor triggers. We further compare the frequency-domain energy distributions of different triggers and evaluate their exposure risk under the Spectral Signatures detection method. Figure~\ref{fig:Frequency-Domain-Features} shows comparisons of detection rates, dominant frequency bands, and PSNR values.

The results indicate that traditional triggers exhibit highly concentrated energy distributions in the frequency domain, leading to fewer dominant frequency bands and higher detection rates. In contrast, fractal triggers exhibit more dispersed broadband distributions, resulting in significantly lower detection rates, approaching the level of random noise. However, the PSNR of fractal triggers remains substantially higher than that of random noise, indicating that they retain structured signal strength necessary for reliable backdoor activation while maintaining frequency-domain stealthiness.

\begin{figure*}[htbp]
	\centering
	\includegraphics[width=0.7\linewidth]{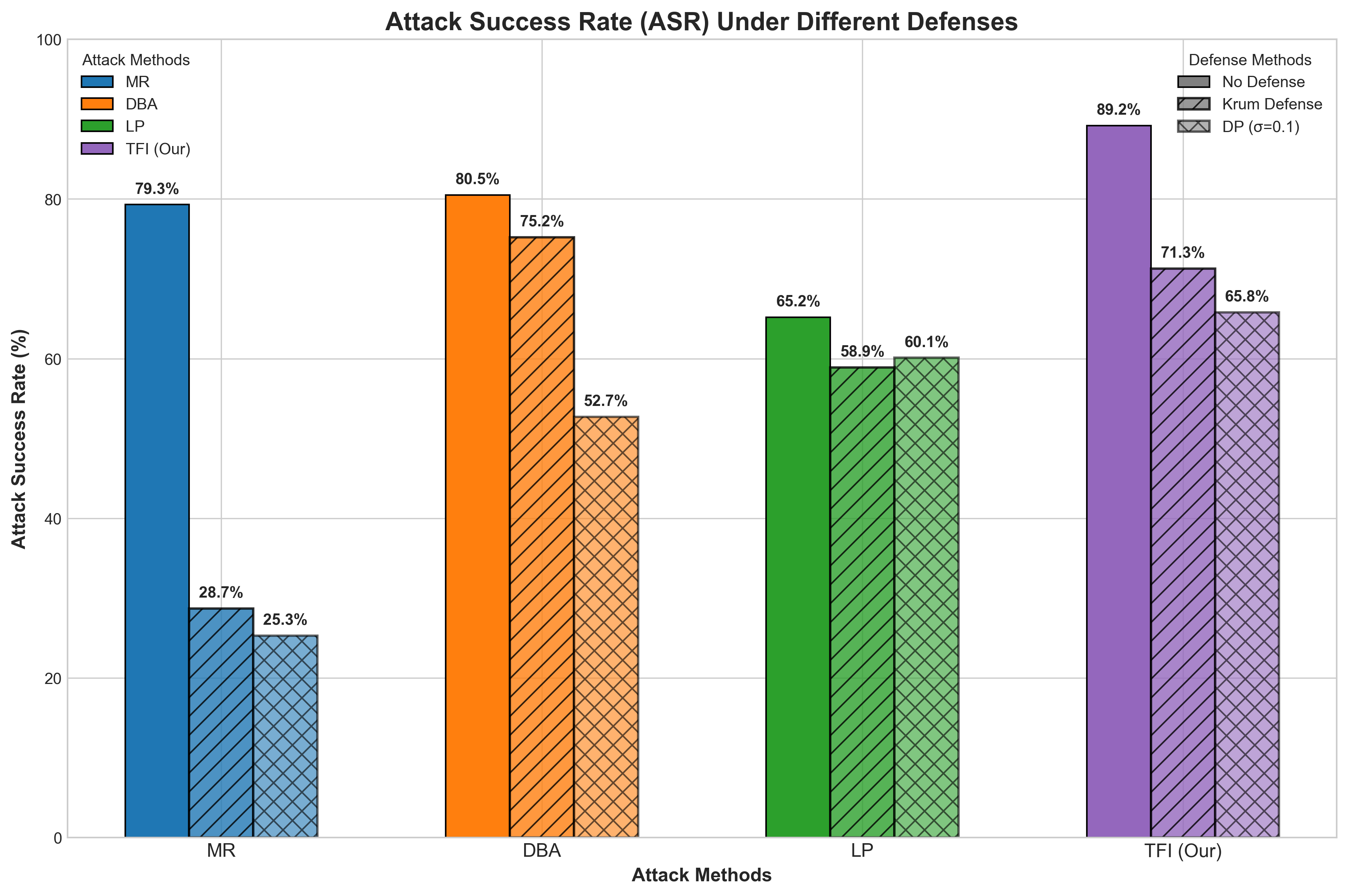}
	\caption{\label{fig:Attack-Success-Rate}Attack success rate under different defense mechanisms}
\end{figure*}

\textbf{Survivability under robust aggregation and differential privacy.} While stealthiness reduces the probability of detection, perturbation survivability ultimately depends on the defense mechanisms employed by the federated learning system. We further analyze the attack retention of TFI under typical robust aggregation algorithms and differential privacy mechanisms. As shown in Fig.~\ref{fig:Attack-Success-Rate}, TFI retains a higher proportion of attack effectiveness under Krum defense, whereas other methods experience significant drops in ASR. This indicates that TFI-generated updates are statistically closer to benign updates and are therefore less likely to be filtered by robust aggregation.

\textbf{Effect of differential privacy noise intensity.} We further analyze the effect of differential privacy noise strength on attack survivability. Table~\ref{tab:The attack success rate of TFI under different differential privacy noise intensities} reports ASR values under different DP noise levels (CIFAR-10, ResNet-18). The results show that under Krum defense, MR and LP suffer substantial performance degradation, while TFI retains a higher fraction of attack effectiveness. Under differential privacy with $\sigma=0.1$, the ASR degradation of TFI remains relatively limited. These results further confirm that TFI-generated updates are more consistent with benign updates and can persist under defense mechanisms.

\begin{table}
	\begin{tabular}{|c|c|c|}
		\hline
		DP noise level $\sigma$ & ASR & ASR retention\tabularnewline
		\hline
		\hline
		No defense & 89.2\% & 100\%\tabularnewline
		\hline
		0.05 & 70.1\% & 78.5\%\tabularnewline
		\hline
		0.10 & 58.7\% & 65.8\%\tabularnewline
		\hline
		0.20 & 37.8\% & 42.3\%\tabularnewline
		\hline
	\end{tabular}
	\caption{\label{tab:The attack success rate of TFI under different differential privacy noise intensities}Attack success rate of TFI under different differential privacy noise intensities}
\end{table}

\subsection{Structural Compatibility and Attack Efficiency}

This subsection quantitatively analyzes the relationship between structural compatibility (SCC) and attack success rate (ASR) to verify whether the impact of model structure on fractal perturbation propagation and retention is reflected in attack efficiency.

\begin{figure*}[htbp]
	\centering
	\includegraphics[width=0.7\linewidth]{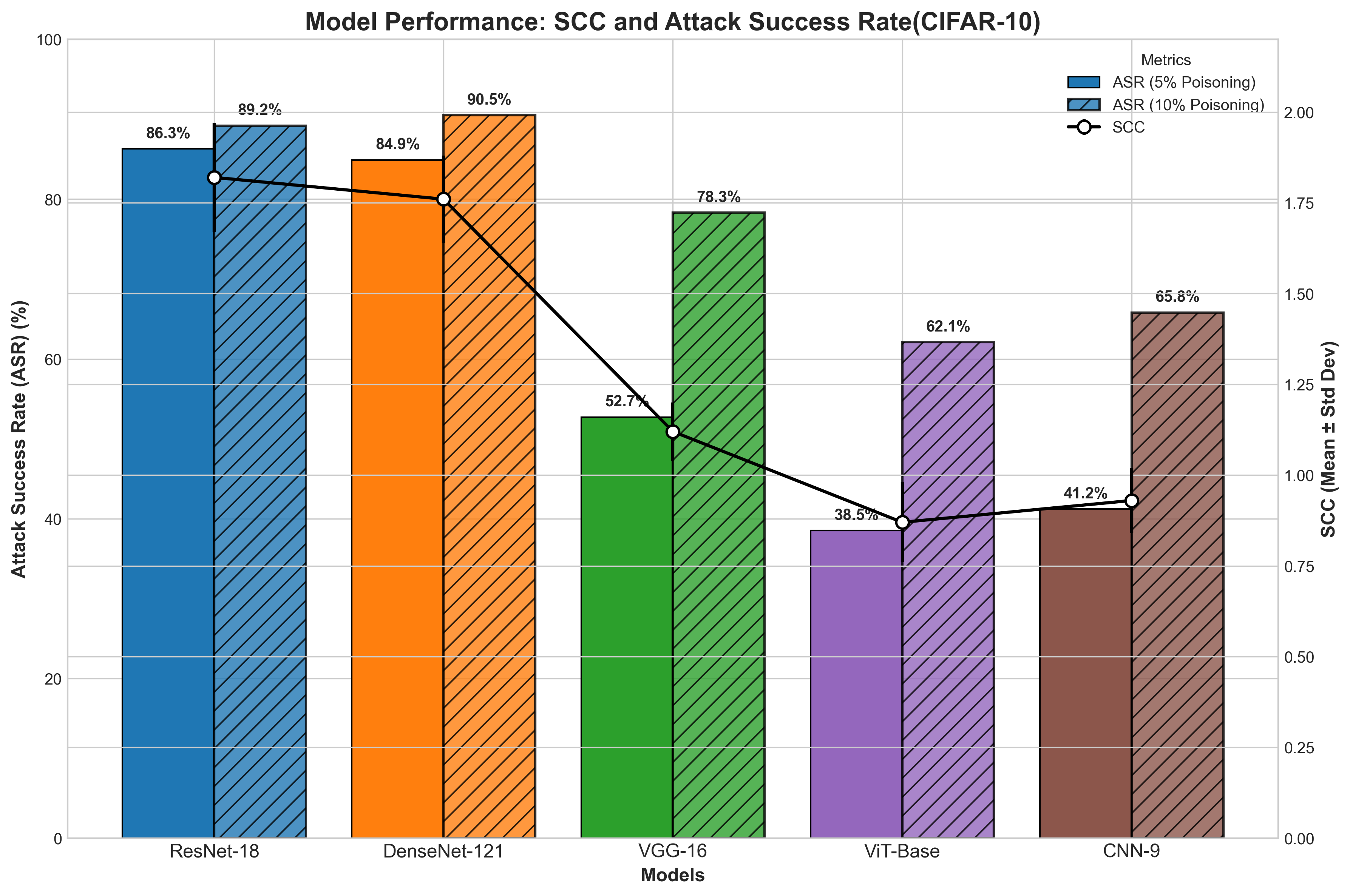}
	\caption{\label{fig:Model-Performance:-SCC}Structural compatibility and attack success rate on CIFAR-10}
\end{figure*}

We measure SCC for different model architectures and compare ASR under identical poisoning ratios. Figure~\ref{fig:Model-Performance:-SCC} shows SCC values and corresponding ASR trends under 5\% and 10\% poisoning ratios. The results reveal significant differences in model responses to fractal perturbations. Models with multi-path feature fusion mechanisms exhibit higher SCC and maintain high ASR even at low poisoning ratios. In contrast, sequential convolutional networks, self-attention models, and shallow CNNs exhibit lower SCC, with ASR dropping more rapidly and becoming more sensitive to poisoning ratio. We further compute the Pearson correlation coefficient between SCC and ASR across model architectures. On CIFAR-10, the correlation coefficient reaches 0.91, indicating a strong positive correlation. These results suggest that SCC not only distinguishes relative structural friendliness to fractal perturbations but also serves as an effective predictor of attack performance.

\begin{figure*}[htbp]
	\centering
	\includegraphics[width=0.7\linewidth]{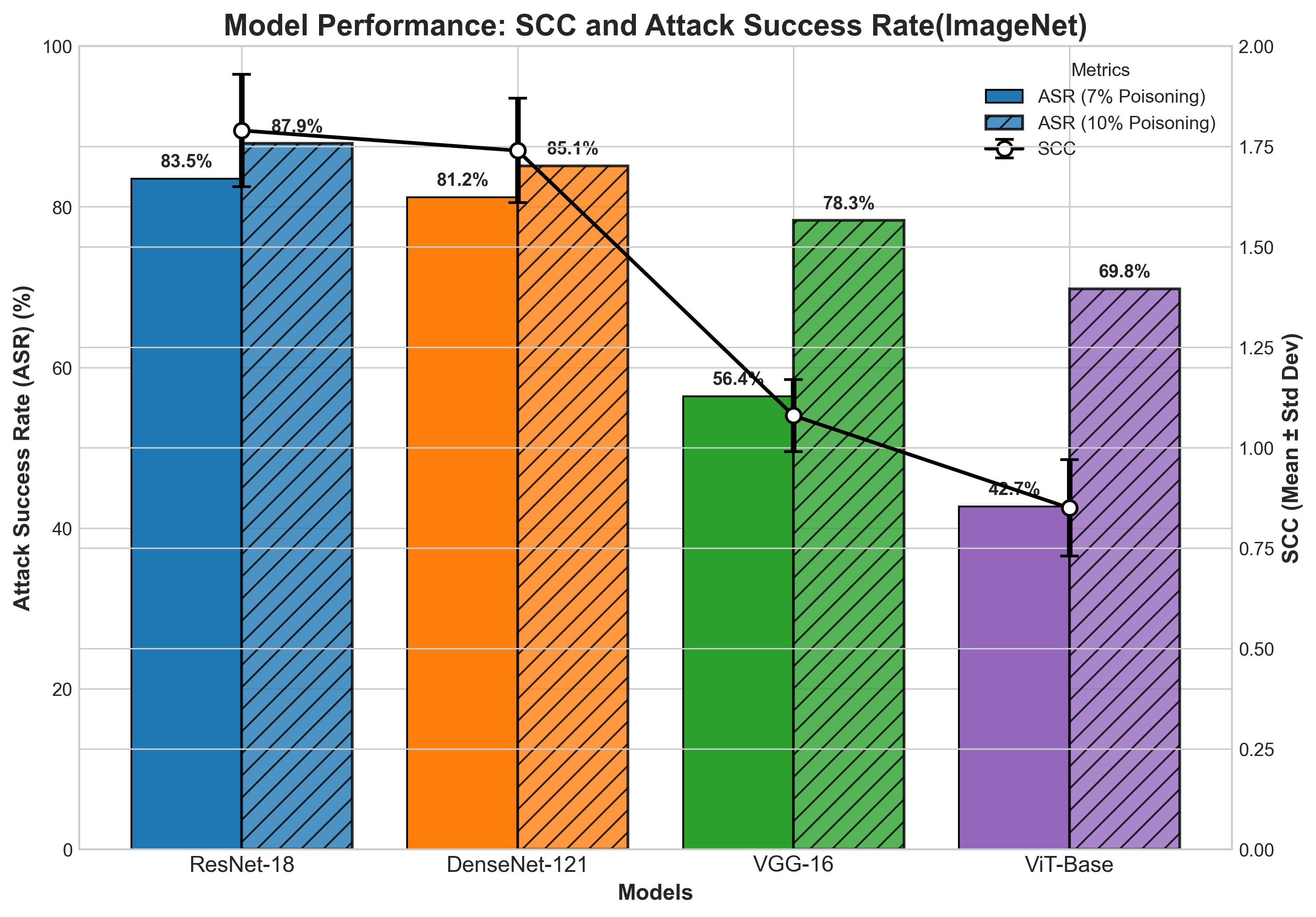}
	\caption{\label{fig:Model-Performance:-SCC-1}Structural compatibility and attack success rate on ImageNet-100}
\end{figure*}

We repeat the same analysis on ImageNet-100 to verify robustness under large-scale tasks. Figure~\ref{fig:Model-Performance:-SCC-1} shows SCC values and ASR trends under 7\% and 10\% poisoning ratios. The consistent relationship between SCC and ASR persists even under higher data and model complexity. Multi-path convolutional architectures achieve high ASR at low poisoning ratios, whereas architectures with lower SCC require significantly higher poisoning ratios to achieve comparable performance, further validating the generalization of the structure-aware analysis framework.

When model structures fail to provide effective propagation paths for fractal perturbations, attack performance degrades in a predictable manner. This degradation behavior is consistent with the theoretical analysis of structural compatibility thresholds, confirming that the advantage of TFI is conditional on explicit structural properties rather than unconditional.

\subsection{Minimum Poisoning Ratio under Fixed ASR}

Having established the correlation between SCC and ASR, we further analyze attack cost by examining the minimum poisoning ratio required to achieve a fixed target ASR under different model architectures.

\begin{figure*}[htbp]
	\centering
	\includegraphics[width=0.7\linewidth]{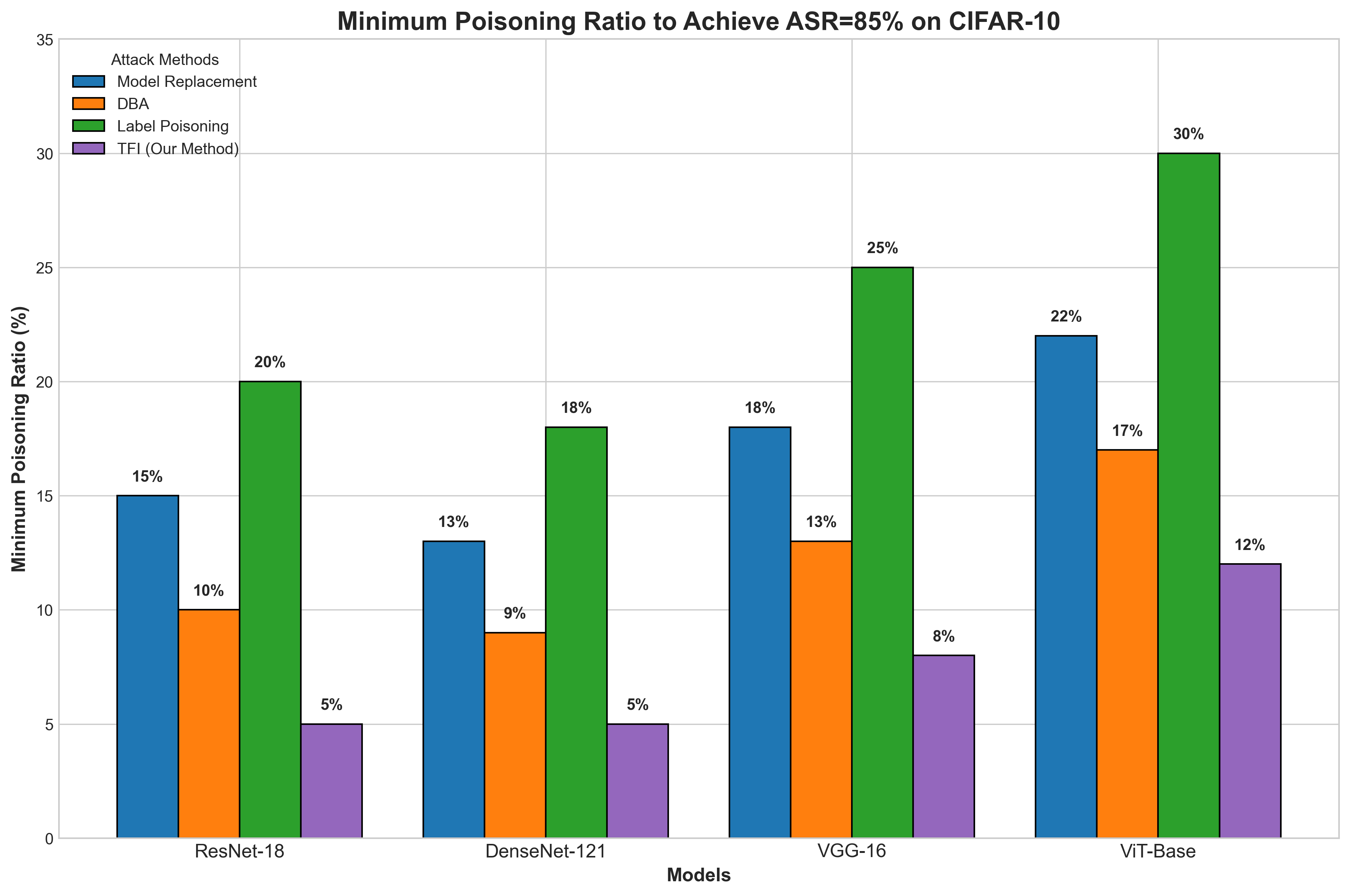}
	\caption{\label{fig:Different model structures on CIFAR-10 reach the minimum poisoning ratio required for ASR.}Minimum poisoning ratio required to reach ASR=85\% on CIFAR-10}
\end{figure*}

Specifically, we fix the target ASR at 85\% and gradually increase the poisoning ratio until the attack success rate stabilizes at this threshold. This process is conducted across different model architectures and attack methods. Figure~\ref{fig:Different model structures on CIFAR-10 reach the minimum poisoning ratio required for ASR.} shows the minimum poisoning ratios required to reach ASR=85\% on CIFAR-10.

Clear structural dependence is observed. In multi-path architectures such as ResNet-18 and DenseNet-121, TFI requires only 5\% poisoning to reach the target ASR, significantly lower than other attack methods. In contrast, in architectures such as VGG-16 and ViT-Base, the poisoning ratio required to reach the same ASR increases substantially. On ViT-Base, TFI requires approximately 12\% poisoning to reach ASR=85\%, indicating a reduced relative advantage.

\begin{figure*}[htbp]
	\centering
	\includegraphics[width=0.7\linewidth]{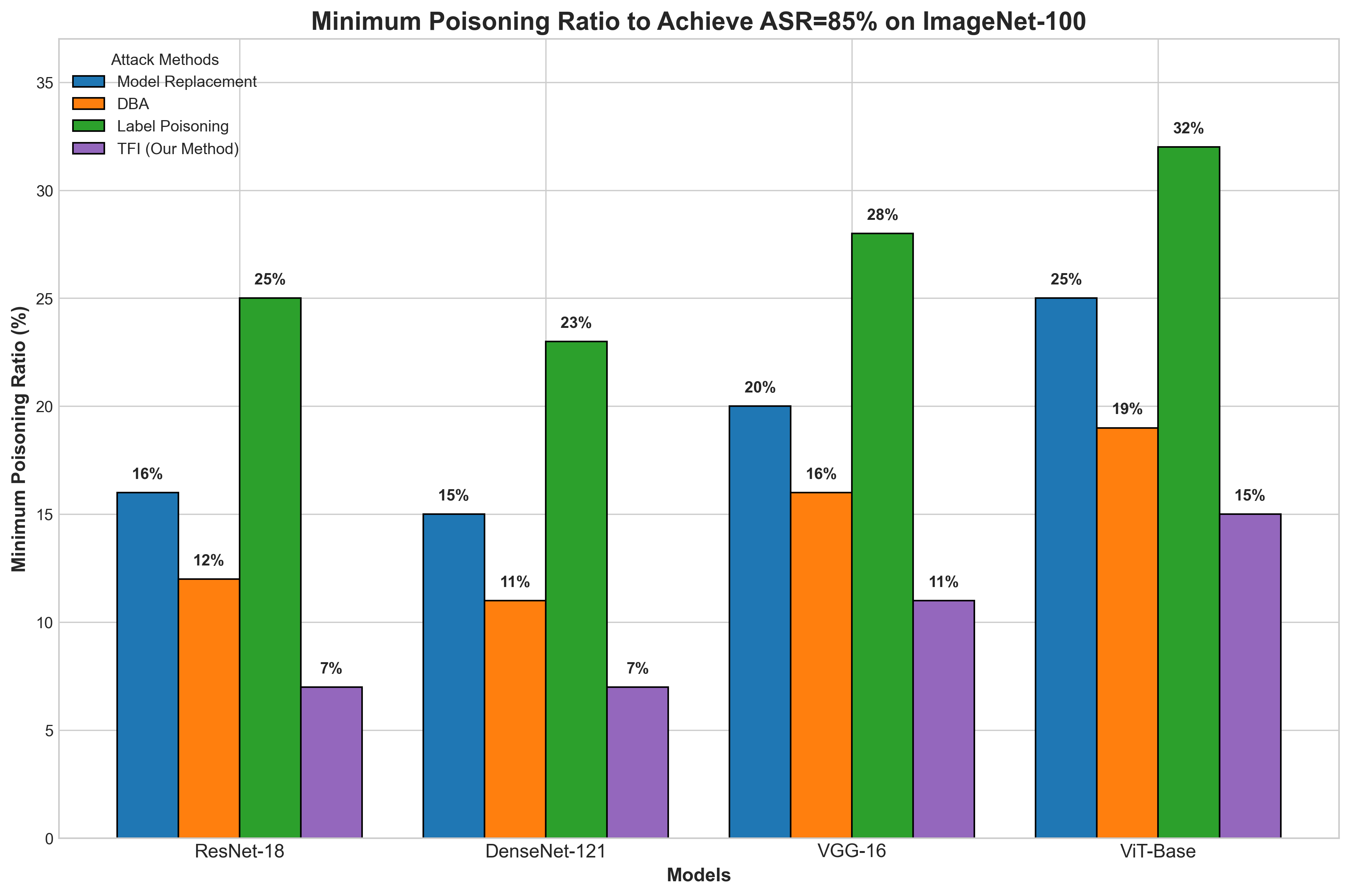}
	\caption{\label{fig:Different model structures on ImageNet-100 reach the minimum poisoning ratio required for ASR.}Minimum poisoning ratio required to reach ASR=85\% on ImageNet-100}
\end{figure*}

We repeat the same experiment on ImageNet-100. Figure~\ref{fig:Different model structures on ImageNet-100 reach the minimum poisoning ratio required for ASR.} presents the corresponding results.

The results indicate that increasing data scale and model complexity does not eliminate structural dependence. Architectures with residual and dense connections achieve the target ASR under substantially lower poisoning ratios, whereas sequential and self-attention-based models require significantly higher attack budgets. These findings confirm that structural compatibility directly determines attack cost under fixed performance targets.

\subsection{Ablation Study}

To further clarify the role of each design component in TFI and verify consistency with theoretical analysis, we conduct systematic ablation experiments by removing key modules individually. The objective is not merely to compare performance, but to identify which mechanisms fundamentally contribute to TFI's advantage and whether they correspond to the core assumptions of the structure-aware framework. All ablation experiments are conducted on CIFAR-10 with ResNet-18 under a fixed poisoning ratio of 5\%. Except for the removed components, all other settings remain identical to the full TFI method. Table~\ref{tab:Comparison of attack performance of TFI under different ablation configurations (CIFAR-10, ResNet-18, 5=000025 poisoning)} summarizes ASR, MTA, anomaly detection rate, and update similarity for different ablation configurations.

\begin{table*}
	\begin{tabular}{|c|c|c|c|c|c|}
		\hline
		Configuration & ASR & MTA & Detection rate & Update similarity & Retention\tabularnewline
		\hline
		\hline
		TFI (full) & 89.2\% & 84.3\% & 18.5\% & 0.87 & 100\%\tabularnewline
		\hline
		w/o SCC-aware client selection & 68.3\% & 85.2\% & 35.7\% & 0.79 & 76.6\%\tabularnewline
		\hline
		w/o fractal perturbation & 72.5\% & 83.8\% & 62.3\% & 0.68 & 81.3\%\tabularnewline
		\hline
		w/o temporal coordination & 75.8\% & 84.1\% & 45.2\% & 0.73 & 85.0\%\tabularnewline
		\hline
		w/o dynamic strength control & 81.6\% & 84.5\% & 28.9\% & 0.82 & 91.5\%\tabularnewline
		\hline
	\end{tabular}
	\caption{\label{tab:Comparison of attack performance of TFI under different ablation configurations (CIFAR-10, ResNet-18, 5=000025 poisoning)}Comparison of attack performance of TFI under different ablation configurations (CIFAR-10, ResNet-18, 5\% poisoning)}
\end{table*}

The results clearly show that different modules contribute unequally to attack effectiveness and stealthiness, consistent with theoretical predictions. Removing SCC-aware client selection leads to the largest ASR drop (89.2\% to 68.3\%), confirming that prioritizing high-SCC clients is critical under low poisoning budgets. Replacing fractal perturbations with static triggers results in increased detection rates and reduced update similarity, indicating that the core role of fractal perturbations lies in enhancing statistical stealthiness rather than merely increasing attack strength. Removing temporal coordination causes moderate ASR degradation and higher detection rates, consistent with its role in avoiding concentrated exposure within individual rounds. Removing dynamic strength control has a smaller impact on ASR but still degrades stealthiness, suggesting it serves as an optimization mechanism rather than a necessary condition for attack success.

Overall, these ablation results demonstrate that TFI's advantage arises from the coordinated effect of multiple structure-aware mechanisms. SCC-aware client selection and fractal perturbations constitute the core prerequisites for attack success, while temporal coordination and strength control primarily enhance stealthiness and stability. Importantly, when any key mechanism is removed, attack degradation follows predictable patterns aligned with its functional role in the theoretical framework.

\subsection{Limitations and Discussion}

The experimental results show that TFI can achieve stable backdoor injection under low poisoning ratios in multi-path architectures and moderate defense settings. However, the method does not succeed unconditionally across all federated learning scenarios. The effectiveness of structure-aware fractal injection fundamentally depends on whether perturbation signals can be continuously amplified, accumulated across rounds, and retained during aggregation.

When model architectures fail to provide low-attenuation propagation paths for structured perturbations, attack effectiveness degrades significantly. This phenomenon is particularly evident in architectures lacking cross-layer shortcuts or exhibiting highly dispersed representations, consistent with the structural compatibility analysis in Section~3. Similarly, attack success depends on maintaining statistical consistency of perturbations across training rounds; once the spectral structure or temporal coherence of fractal perturbations is disrupted, their accumulation in the global model diminishes substantially.

At the system level, federated aggregation mechanisms and noise injection further constrain attack feasibility. Robust aggregation and differential privacy introduce effective noise during aggregation, and when perturbation strength falls below corresponding thresholds, its impact is weakened or completely submerged. In addition, the inherent random client participation in federated learning may interrupt continuous attack injection over time, especially under low participation rates, amplifying the risk of attack failure.

These limitations are not independent failure modes, but rather manifestations of a single underlying constraint: whether the effective signal-to-noise ratio of fractal perturbations in federated learning remains within a feasible range under the joint effects of model structure, training dynamics, and aggregation mechanisms. From this perspective, TFI represents a structure-dependent attack paradigm whose success and failure are both predictable and interpretable. From a defensive standpoint, weakening perturbation amplification paths in model architectures, disrupting cross-round statistical consistency, or introducing sufficient system noise during aggregation can all effectively suppress structure-aware backdoor attacks. This analysis provides clear guidance for future defense research at the levels of model architecture, training dynamics, and aggregation mechanisms.

\section{Conclusion}

To address the issues of high dependency on poisoned samples and insufficient stealth in distributed backdoor attacks within federated learning, this paper proposes an efficient and stealthy attack framework named FDBA (Fine-grained Distributed Backdoor Attack). By leveraging fine-grained trigger generation—based on precise Canny edge structures and strategically placed Laplacian noise with RGB channel decomposition—combined with targeted contrastive embedding optimization, FDBA achieves impressive performance even with a poisoning rate below 5\%, reaching 94.5\% ASR on CIFAR-10 and 93.8\% on ImageNet, outperforming traditional methods by over 8.1\%.

Experimental results demonstrate that FDBA significantly enhances visual stealthiness (PSNR $>$ 38 dB, SSIM $>$ 0.98) and anti-detection robustness (anomaly detection rate $<$ 6.2\%) compared to existing approaches. In Non-IID scenarios, FDBA exhibits 44\% less degradation in attack success rate, and successfully circumvents mainstream defenses such as Krum and FoolsGold.

Ablation studies validate the indispensable synergy among dynamic trigger generation, contrastive learning, and random projection hashing. The removal of any component results in a substantial ASR drop of 23.6\% to 26.4\%. This study reveals the severe threat posed by low-poisoning backdoor attacks to federated learning systems and offers new insights for designing adaptive and dynamic defense strategies. Future work will focus on cross-modal federated scenarios, aiming to establish robust attack-defense game theories and adaptive defense mechanisms.


\printcredits

\bibliographystyle{cas-model2-names}


\bibliography{references}

\end{document}